\documentclass[letterpaper]{article} 
\usepackage{aaai18}  
\usepackage{times}  
\usepackage{helvet}  
\usepackage{courier}  
\usepackage{url}  
\usepackage{graphicx}  
\usepackage{color} 
\usepackage{mathtools} 
\usepackage{amsthm} 
\usepackage{xspace} 
\usepackage{array} 
\usepackage{amsfonts}
\usepackage{subcaption}
\usepackage[vlined, ruled]{algorithm2e}

\def\year{2019}\relax

\usepackage{xcolor}
\usepackage{amssymb}
\usepackage{pifont}

\begin{document}
%

\title{Artificial Intelligence: Powering Human Exploration of the Moon and Mars}
\author{Jeremy D. Frank \\ NASA Ames Research Center \\ jeremy.d.frank@nasa.gov \\}

\nocopyright
\maketitle



\section{Introduction}
\begin{figure*}
\centerline{\includegraphics[width=6.5in]{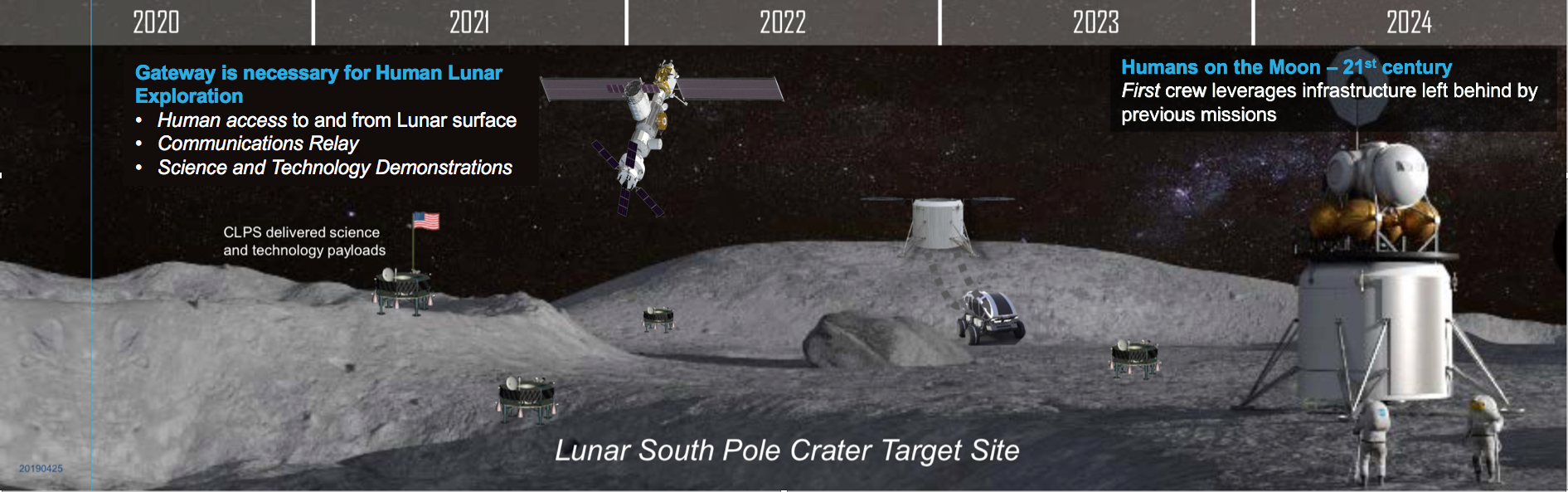}}
\caption{ \label{Phase1Lunar}  The first phase of Artemis will lead to NASA returning humans to the Moon by 2024.}
\end{figure*}

NASA is committed to landing American astronauts, including the first woman and the next man, on the Moon by 2024. Through the agency's Artemis lunar exploration program, we will use innovative new technologies and systems to explore more of the Moon than ever before. 
In  support of this vision, NASA plans to construct Gateway, a habitable spacecraft
\cite{kn:Gateway}, 
in the vicinity of the Moon.  Gateway consists of a Habitat, Airlock, Power and Propulsion Element (PPE), and Logistics module. 
The PPE provides orbital maintenance, attitude control, communications with Earth, space-to-space communications, and radio frequency relay capability in support of extravehicular
activity (EVA) communications.  The Habitat provides habitable volume and short-duration life support functions for crew in cislunar space, docking ports, attach points for external robotics, external science and technology payloads or rendezvous sensors, and accommodations for crew exercise, science/utilization and stowage.
The Airlock provides capability to enable astronaut EVAs as well as the potential to accommodate docking of additional elements,
observation ports, or a science utilization airlock.  A Logistics module delivers cargo to the Gateway.  
Figure \ref{Phase1Lunar} shows the progression of missions that will ultimately lead to the return to the Moon.
NASA will collaborate with commercial and international partners to establish sustainable missions by 2028. And then we will use what we learn on and around the Moon to take the next giant leap - sending astronauts to Mars.



For over 60 years, NASA's crewed missions have been confined to the Earth-Moon system, where speed-of-light communications delays between crew and ground are practically nonexistent.  The close proximity of the crew to the Earth has enabled NASA to operate human space missions primarily from the Mission Control Center on Earth. This ``ground-centered" mode of operations, with a large, ground-based support team, has had several advantages: the on-board crew could be smaller, the vehicles could be simpler and lighter, and the mission performed for a lower cost.  Despite these advantages, NASA has invested in numerous Artificial Intelligence (AI) techniques to make Human Spaceflight Mission Control more efficient and effective \cite{kn:OpsTech},
as well as to automate some functions onboard the International Space Station \cite{kn:HAL}.

Gateway, however, will feature four crew instead of 6 (or more), and will be occupied for only a few months out of the year. 
Small crews cannot take on all Gateway functions performed by ground today, and so vehicles must be more automated to reduce the crew workload for such missions.  In addition, both near-term and future missions will feature significant periods when crew is not present;  
Gateway shall provide for autonomous operations for up to 21 days, independent of ground communications, when crew are not present.
A thorough assessment of human spaceflight dormancy and autonomy requirements for a future Mars mission \cite{kn:Dormancy} describes the wide range of challenges to be faced.  
The need for autonomy is reflected in the Gateway concept of operations 
\cite{kn:Gateway} 
and its requirements, as well as requirements for the first of its components, the PPE
\cite{kn:PPE}.
Gateway may benefit from robotic assistants inside the spacecraft, and robots may perform some functions while astronauts are not present.
Finally, future missions to Mars will require crew and ground to operate independently from Earth-based Mission Control.
Due to the mission-critical nature of human spaceflight, autonomy technology must be robust and resilient, both to changes in environment, faults and failures, and also the somewhat unpredictable nature of human-machine collaboration.
Maturing autonomy technologies for use on Gateway, and other Artemis elements, will require early adoption to choose the best solutions and determine best engineering practices, and stress the interaction between multiple technologies as well as human-machine interaction.

Artificial Intelligence (AI) is a growing field of computational science techniques designed to mimic functions performed by people.  Advancements in autonomy will depend on a portfolio of AI
technologies.  Automated planning and scheduling is a venerable field of study in AI, and is needed for a variety of mission planning functions.  Plan execution technology is less well studied, but important for autonomy and robotics.  Specialized forms of automated reasoning and machine learning are key technologies to enable fault management. 
Over the past decade, the NASA Autonomous Systems and Operations (ASO) project has developed and demonstrated numerous autonomy enabling technologies employing AI techniques.  Our work has employed AI in three distinct ways to enable autonomous mission operations capabilities.
{\em Crew Autonomy} gives astronauts tools to assist in the performance of each of these mission operations functions.  {\em Vehicle System Management} uses AI techniques to turn the astronaut's spacecraft into a robot, allowing it to operate when astronauts are not present, or to reduce astronaut workload.  AI technology also enables {\em Autonomous Robots} as crew assistants or proxies when the crew are not present.
When these capabilities are used to enable astronauts to operate autonomously, they must be integrated with user interfaces, introducing numerous human factors considerations; when these capabilities are used
to enable vehicle system management, they must be integrated with flight software, and run on embedded processors under the control of real-time operating systems.

We first describe human spaceflight mission operations capabilities.  The remainder of the paper will describe the ASO project, and the development and demonstration performed by ASO since 2011.  
We will describe the AI techniques behind each of these demonstrations, which include a variety of symbolic automated reasoning and machine learning based approaches.  Finally, we  conclude with an assessment of future development needs for AI to enable NASA's future Exploration missions.

\section{Human Spaceflight Mission Operations}
NASA's human spaceflight mission operations functions are performed by flight controllers, both before and during missions.  For the purposes of this work, we focus on functions performed during the mission, and break these functions into Monitoring, Planning, Execution and Fault Management.  

Monitoring addresses the following question: {\em What is the state of the spacecraft?}
Monitoring requires processing and abstracting data from sensors.
Flight controllers group the resulting information into displays, both physical and logical, that contain information specific to major spacecraft system and phases of mission.
For instance, power systems data is monitored by one operator, while attitude and orientation information is monitored by a different operator.
Monitoring ensures plans are being executed as expected, and the spacecraft state and the state of all relevant systems is known.

Planning addresses the question: {\em What is the spacecraft doing, and when?}
Plans are created to achieve specific objectives. 
For most spacecraft, plans are created days to weeks (or even months) ahead of time.
Plans are often created for major spacecraft subsystems separately, then integrated.  For example, a plan to orient the solar arrays of the International Space Station (ISS) is created independently, by one team of operators, from the plan for crew activities, which is created by a second team.  These plans must be integrated, and discrepancies resolved, prior to execution. 
Unexpected events or faults may require replanning on shorter time scales, which is discussed in subsequent paragraphs.

Plans are no good unless they are executed.  Execution addresses the question: {\em What is the next activity to perform?  When it is performed, what was the outcome?}
Execution ultimately involves issuing commands to spacecraft subsystems and ensuring results are as expected.   Execution for human spacecraft such as the International Space Station also requires ensuring astronauts have performed their activities as planned.  Execution ultimately is driven by planning, and results in the need to monitor spacecraft systems to ensure the results are as expected. 
If unexpected events occur, replanning may be needed. 

Fault management addresses the question: {\em Is something wrong?  If so, what has gone wrong, and what are the impacts?}
Fault management involves the detection and isolation of faults.  Monitoring may provide the first indication of faults or anomalies; anomalies may be the precursors of faults.
Faults often come not as single spies, but in battalions; in such cases, it is important to determine the root cause of many faults.
Fault management also involves determining the consequences of faults.
Recovering from or mitigating faults also involves replanning. 

\section{Autonomous Systems and Operations}
NASA's Autonomous Systems and Operations (ASO) project\footnote{Previously called the Autonomous Mission Operations project.} has extensively demonstrated the use of AI techniques to help monitor, plan, execute, and manage faults for future human space missions.
The ASO project has successfully demonstrated autonomy capabilities in all three previously described main areas:  
Crew Autonomy, Vehicle Systems Management, and Autonomous Robotics.  We describe ASO's activity in each of these areas below, emphasizing the specific AI techniques in each demonstration.  A quantitative summary of AI techniques used for each demonstration also appears in the summary (see Figure \ref{UsesWithMetrics}).

\subsection{Crew Autonomy}
In \cite{kn:AMO}, ASO demonstrated the impact of {\em planning and fault management} on mission operations in the presence of time delay.
Mission operation scenarios were designed for NASA's Deep Space Habitat (DSH), an analog spacecraft habitat, covering a range of activities including nominal objectives, system failures, and crew medical emergencies. The scenarios were simulated at time delay values representative of Lunar (1.2-5 sec), Near Earth Object (NEO) (50 sec) and Mars (300 sec) missions. Each combination of operational scenario and time delay was tested in a Baseline configuration, designed to reflect present-day operations of the ISS, and a Mitigation configuration in which AI-based planning and fault management tools, information displays, and crew-ground communications protocols were employed to assist both crews and Flight Control Team (FCT) members with mitigating the long-delay conditions. Cognitive workload of both crewmembers and FCT members generally increased along with increasing time delay. 
Autonomy enabling technology decreased the workload of both flight controllers and crew, and decreased the difficulty of coordinating activities. 

\begin{figure}[t]
\centerline{\includegraphics[width=3in]{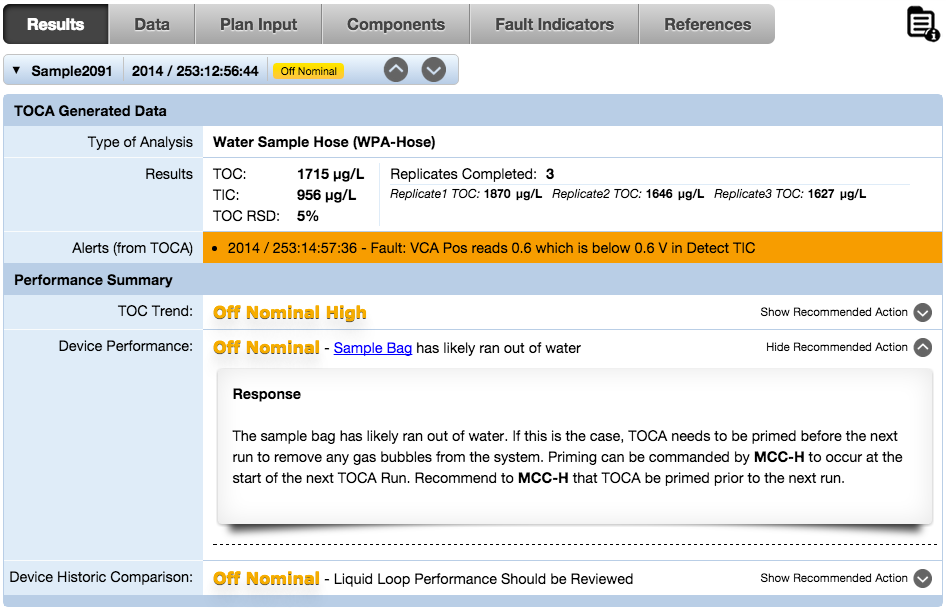}}
\caption{ \label{TOCAUI}  User interface for AMO TOCA SSC demonstration, showing a TOCA error detected by fault management displayed for ISS crew.}
\end{figure}

The AMO TOCA SSC demonstration \cite{kn:TOCA} demonstrated {\em planning and fault management} of two spacecraft systems onboard the International Space Station (ISS).
Astronauts managed the Total Organic Carbon Analyzer (TOCA), a water quality analyzer, and Station Support Computers (SSC) laptops, which are non-critical crew computer systems. These systems were selected as representative of systems a future crew may need to operate autonomously during a deep space mission. ISS astronauts autonomously operated these systems (see Figure \ref{TOCAUI}), taking on mission operations functions traditionally performed by support teams on the ground, using new software tools that provide decision support algorithms for planning, monitoring and fault management, hardware schematics, system briefs, and data displays that are normally unavailable to the crew. 
Managing TOCA required monitoring 22 data items for detecting anomalies, generating plans of up to 6 activities (of 12 possible) subject to 10 constraints, and detecting 70 faults.
Monitoring SSCs required monitoring 161 data items for faults. 
The experiment lasted seven months, during which ISS crews managed TOCA and SSCs on 22 occasions. The AMO software processed data from TOCA and SSCs continuously during this seven month period. 
The combined performance of the software and crew achieved a 88\% success rate on managing TOCA activity, the system for which ground-truth was available.


The AMO EXPRESS and AMO EXPRESS 2.0 {\em plan execution} demonstrations automated activation and de-activation of a facility experiment rack with a single operator action, described in \cite{kn:EXPRESS}. The auto-procedures perform combined ``core" (e.g. power, thermal, life support) and payload commanding with embedded fault detection and recovery, essentially performing the specific operations responsibilities for three flight controllers. In addition, the auto-procedures were integrated with a Web-based procedure execution monitoring system. 
The procedure executes 59 individual commands, and monitored 236 distinct data items.
A first demonstration showed that the facility could be powered up and configured automatically from the ground; a second demonstration showed that the the task could be performed and monitored by ISS astronauts.  In both demonstrations, ISS flight controllers scheduled the activity, which astronauts performed at the designated time; this demonstration is being extended to require {\em planning,} i.e. astronauts will also decide when to schedule the activity based on a variety of constraints.

\begin{figure}[h]
\centerline{\includegraphics[width=3in]{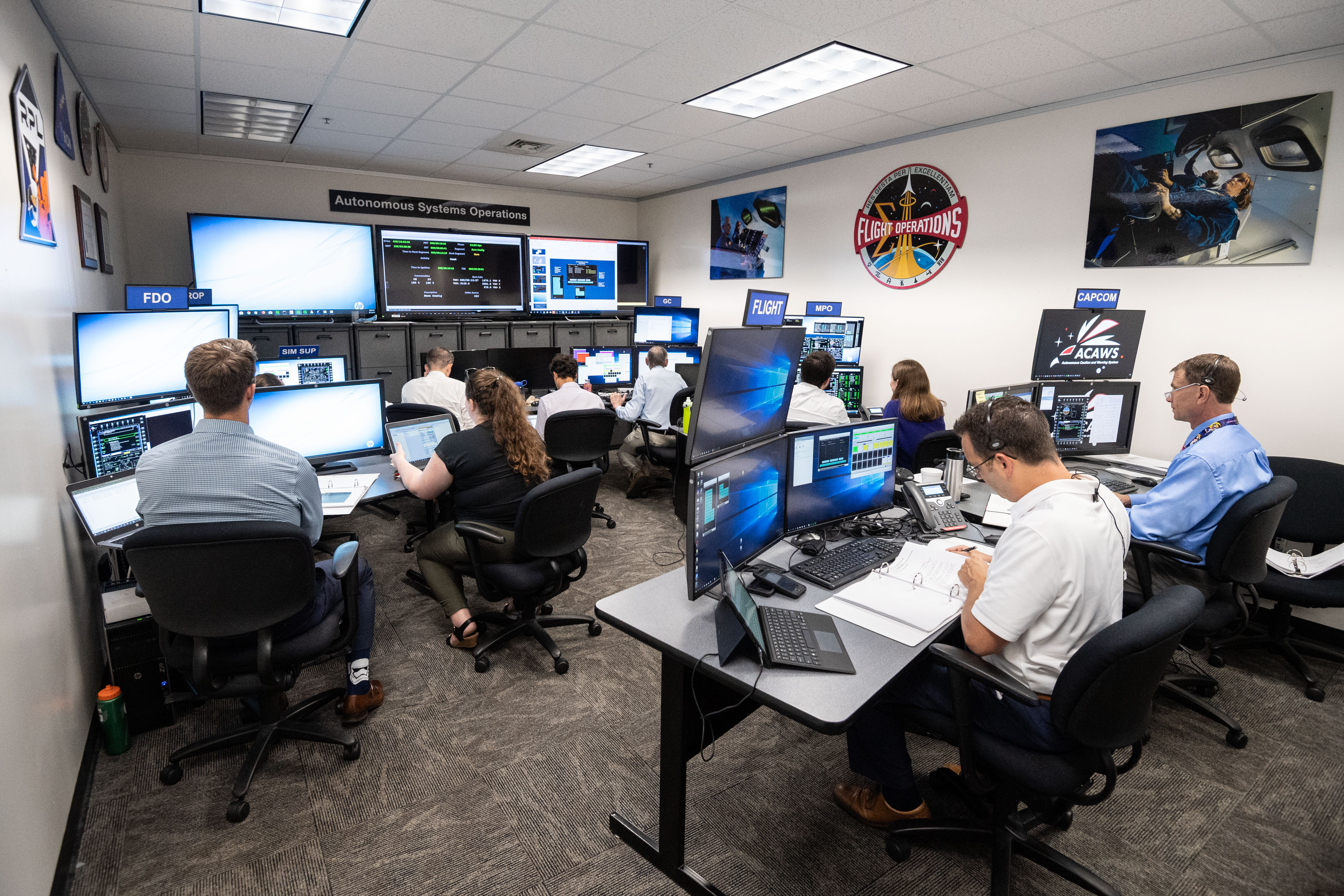}}
\caption{ \label{ACAWS} Simulated Control Center used to evaluate ACAWS.  Both Flight Controllers and crew have access to ACAWS output.}
\end{figure}

ASO developed and tested an automated model-based {\em fault management}
called Advanced Caution and Warning (ACAWS). ACAWS was tested on Orion Exploration Flight Test 1 (EFT-1) data from prelaunch to post- landing, as
described in \cite{kn:AaBaVaMo}. 
The system was tested extensively using nominal test data from the Orion Program and combined with fault signature data developed by the ACAWS test team and supported by NASA Flight Operations Directorate. Nominal data was recorded when Orion test data was transmitted to the Mission Control Center, either from the Orion EFT-1 spacecraft during vehicle checkout, or from Orion test and simulation systems.
The ACAWS system was configured to detect and display approximately 3500 failure modes, ranging from sensor faults to major system component failures, monitoring approximately 2500 unique telemetry elements. The system executed on a Linux laptop at a 1 Hz execution rate.
Flight controller evaluation indicated that the information provided by ACAWS would be an effective aid in rapidly understanding failures and responding more quickly and accurately with adequate training and familiarity.  

\subsection{Vehicle System Management}
ASO integrated {\em planning, plan execution and fault management} technologies into a Vehicle System Manager (VSM) able to autonomously operate a dormant (uncrewed) space habitat,
described in \cite{kn:VSM}. These technologies included a fault-tolerant avionics architecture, novel spacecraft power system and power system controller, and autonomy software to control the habitat.
The power system controller was developed by a partner project \cite{kn:AMPSold}, \cite{kn:AMPS}. 
The demonstration involved simulation of the habitat and multiple spacecraft sub-systems (power storage and distribution, avionics, and air-side life-support). The foundation of the demonstration was `quiescent operations' of a habitat during a 55 minute eclipse period. For this demonstration, the spacecraft power distribution system and air-side life support system were simulated at a high level of fidelity; additional systems were managed, but with lower fidelity operational constraints and system behavior. Operational constraints for real and simulated loads were derived from ISS hardware and future Exploration capable technology. A total of 13 real and simulated loads were used during the test. 
The VSM monitored 208 parameters, was able to detect and respond to 159 faults, and managed plans of 312 steps in the presence of 13 load constraints, 6 power system configuration constraints, and 6 operational constraints (e.g. two systems must be on or off simultaneously, a system cannot remain off for more than 30 minutes, etc).
Evaluation was performed with eight scenarios including both nominal and off- nominal conditions.  We have subsequently scaled the number of loads (to 24), activities, operational constraints (roughly 50) and fault modes.

ASO has subsequently adapted the ACAWS {\em fault management} technology as an autonomous VSM function; 
while Orion may be in Earth proximity, fault management will improve the crew's situational awareness, and may be important in the event of loss of communications.
In \cite{kn:VSMACAWS}, \cite{kn:AaFr} we describe the deployment of ACAWS on an embedded flight computer and a real-time operating system.  The considerably larger size of the ACAWS fault model, and the larger number of parameters to monitor, make this a more useful stressing case than the Habitat-based VSM described above (5000 faults for the current Orion model versus 159 faults for the Habitat model.)

\subsection{Robotics}
ASO demonstrated {\em planning and plan execution} of a dexterous manipulator performing an ISS-inspired biological science scenario
\cite{kn:ASORobot}. The demonstration system was implemented using the Robot Operating System (ROS) for inter-process communication between planning, perception and control, ROSPlan to implement task-level planning and execution, and affordance templates to perform kinematics and continuous path planning.  The test setup utilizes the Sawyer robot with electric parallel gripper manufactured by Rethink Robotics, and a mockup ISS experiment facility with objectives such as manipulating test tubes for a science experiment.
The combined system was used to plan activities such as the inspection of test tubes, simulated activities such as heating and shaking, sending data to ground to await confirmation to proceed, etc.  Execution of the plans can be disrupted by long wait times, switching test tube positions, and other unexpected events.
Plans range in size from 50-70 actions; there are 28 distinct action types, so many actions are repeated in these plans.  A total of 54 predicates describing the robot and experiment rack state are monitored during execution

\section{Artificial Intelligence Enabling Autonomous Mission Operations}
We describe the specific AI applications employed by each ASO demonstration in more detail below.

\subsection{Planning and Scheduling}
The planner used in \cite{kn:TOCA} is custom code designed specifically to generate new TOCA activities based on TOCA operating constraints and the current schedule.
We chose this design due to the simplicity of generating a small set of TOCA activities 2-3 weeks into the future, allowing us to focus on deployment and system integration with other AI components needed to manage TOCA.

\begin{figure}[t]
\centerline{\includegraphics[width=3in]{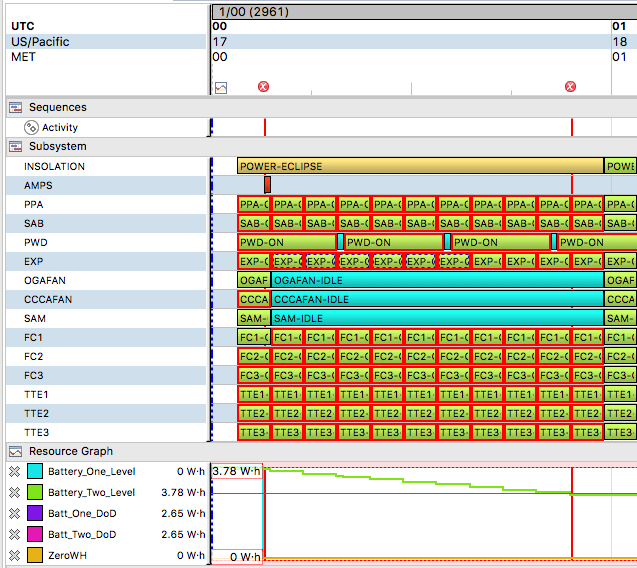}}
\caption{ \label{Powerfault}  VSM planning and replanning in the presence of power systems faults.}
\end{figure}

The VSM planner uses the Solving Constrained Integer Programs (SCIP) Optimization tool \cite{kn:SCIP}, an open-source tool designed to solve constrained optimization problems which combine mixed-integer, and linear programming, and constraint programming methods. We modeled the power loads as jobs to be scheduled. Each job had periodic constraints to model duty cycles for each load, constraints requiring some loads power mode to be synchronized with each other, and constraints on the maximum instantaneous power demand and total energy consumption over the 2-hour plan horizon (see Figure \ref{Powerfault}).  As the mission continued, or when new circumstances (e.g. advancing time, new goals or constraints, unexpected events, or faults) arise, the scheduler generates new plans.  The SCIP solver is re-invoked every 5 minutes as new information becomes available, or when faults occur.

The planner used in our robotics work is POPF \cite{kn:CoCoFoLo}, a model-based planner that employs a declarative action model.  
Planning problems are formulated by defining a Plan Domain Description Language (PDDL) 2.1 domain. For more information about this form of planning, we refer the reader to \cite{kn:TrGhNa}.
Plan actions are mapped to affordance templates, which provide a way of describing an action that an object affords to the robot, such as picking up a test tube. The construct of an affordance template allows multiple methods to achieve different goals for the same objects, e.g. pick an object versus place an object.

\subsection{Execution}
The AMO EXPRESS demonstrations onboard ISS employed Timeliner as an executive.
Timeliner \cite{kn:Timeliner} provides a way to programmatically model human decisions. Its constructs relate directly to the decision processes made by ISS flight controllers each day concerning such things as activating and monitoring facilities and payloads and any other procedural processes. Timeliner has English language constructs that allow someone with little programmer experience to not only follow and understand execution, but to write scripts. 
Timeliner scripts are hierarchically organized.  The top level construct, the bundle, is a place holder for sequences. Within each bundle there has to be at least one sequence, which is the executable code that can start executing autonomously or by ground commanding. 
In addition to the natural language control constructs and the hierarchical nature of bundles and sequences, Timeliner is well integrated with the ISS command and telemetry system, making integration with ISS flight software quite straightforward.

The VSM demonstration employed 
The Plan Execution and Interchange Language (PLEXIL) as an executive. PLEXIL was developed as a collaborative effort by NASA and Carnegie Mellon University, and subsequently released as open-source software \cite{kn:VeJoPaIa}. PLEXIL is a language for representing flexible robust plans intended to be executed in an uncertain and changing environment.  
PLEXIL provides numerous conditions to control the execution of plan elements, including loops and hierarchical decomposition,
and provides internal plan element state tracking that can be used natively as part of control logic.
PLEXIL provides well defined execution semantics with contingency handling which can be formally validated and produces deterministic results given the same sequence of environmental inputs. PLEXIL's Execution Engine (executive) executes stored plans written in the PLEXIL language; it can load new plans during execution.

The Robotics work employs ROSPlan as an executive.
ROSPlan contains a Knowledge Base that manages the evolving state of the world and interprets a model for planning. The deliberative layers consist of a ROSPlan manager (RPM), that commands and monitors the Knowledge Base, and the planning system for generating and executing task plans. The plan dispatcher commands and receives feedback from the robot behavior components, which are themselves composed of sequences of simple actions, and updates the Knowledge Base. 
ROSPlan components were designed with a one to one relationship to the actions in the PDDL domain model. When ROSPlan dispatched an action to the desired component, the component would invoke the behavior required for completing the action. Upon successful completion of the behavior, we would update the ROSPlan Knowledge Base with the effects of the action, using sensor data as verification prior to updating as applicable.

\subsection{Fault Management}
Both our VSM work and our Orion fault management work employ the Testability Engineering and Maintenance System (Real Time) (TEAMS-RT) \cite{kn:TEAMS} software as a foundation.  TEAMS models consist of components, fault modes for each component, a component interconnection model representing transfer of capability (e.g. flow of current or data), and a set of (logical) tests.  Each
test is associated with a set of fault modes it is able to identify, which is a function of the inter-connectivity of components.
Tests can pass, fail or be unknown. TEAMS-RT correlates test points with failure modes. A failed test can implicate one or more failure modes, while a passing test exonerates failures. With sufficient data, a single failure mode can be identified that is responsible for all failed tests, and an unambiguous failure mode is identified. Otherwise, TEAMS determines the smallest set of possible failure modes that could be responsible for failed tests and presents an ambiguity group of possible failures. 
TEAMS on its own performs fault isolation, but requires upstream processing to transform sensor data from systems into test pass-fail inputs.  While TEAMS is commercial software, we created the up-stream code for VSM and Orion fault detection.

The Fault Impacts Reasoner (FIR) determines the resultant impacts of confirmed failures  \cite{kn:AaBaVaMo}. Impacts include the loss of function due to a fault, such as the components that have lost electrical power due to a fault in the electrical system. 
The loss of redundancy due to a fault is also determined. Most critical functions in spacecraft depend on redundancy to assure the availability of the function in spite of failures. Of particular concern is any function that could be lost by a single additional failure, or has become zero-fault tolerant. 
Fault impacts are determined by tracing the inter-connectivity graph from the faulty component, identifying loss of capability along component interconnections.  When there are multiple paths between components, loss of redundancy is also computed.
While FIR's reasoning algorithm was implemented by NASA, it uses the TEAMS model as a foundation.


The TOCA crew autonomy demonstration and the VSM demonstration used the Hybrid Diagnosis Engine (HyDE) \cite{kn:HyDE} for fault detection and isolation.
HyDE  uses hybrid (combined discrete and continuous) models and sensor data from the system being diagnosed to deduce the evolution of the state of the system over time, including changes in state indicative of faults.  HyDE models are state transition diagrams, showing how events change system state; these events can be nominal (e.g. commands and processes) or faults.  When the sensor data are no longer consistent with the nominal mode transitions, HyDE determines what failure mode or modes are now consistent with the data.  While HyDE requires some preprocessing of system sensor data, HyDE can take general inputs (commands, numerical values) from systems, making it more powerful than TEAMS-RT, but more computationally expensive.
 In contrast with TEAMS, HyDE can represent {\em hybrid} systems, i.e. mixes of discrete and continuous quantities; TEAMS, by contrast, can only model discrete systems.

Anomaly detection was performed for TOCA and for batteries during an early VSM demonstration.
Anomaly detection is performed using two linked algorithms.  The Inductive Monitoring System (IMS) \cite{kn:Iv} is used to learn a model of the nominal performance of a system using recorded data gathered from prior runs via clustering.  
%
Once the clusters are created, new instances of system device behavior
can be compared to the previously learned behavior, and the distances $\delta(b,c)$ are computed.  In previous incarnations of IMS, if the new vector $b$ is `too far away' from all clusters, an anomaly is declared.  The Meta-Monitoring System (MMS) generalizes this idea by computing the probability distribution of the cluster distances of the training data; if the new vectors' distance from all clusters is `too unlikely' according to this probability distribution, an anomaly is declared \cite{kn:TOCA}.   


\section{Autonomy Technology Interaction Paradigms}
\begin{figure}[t]
\centerline{\includegraphics[width=3in]{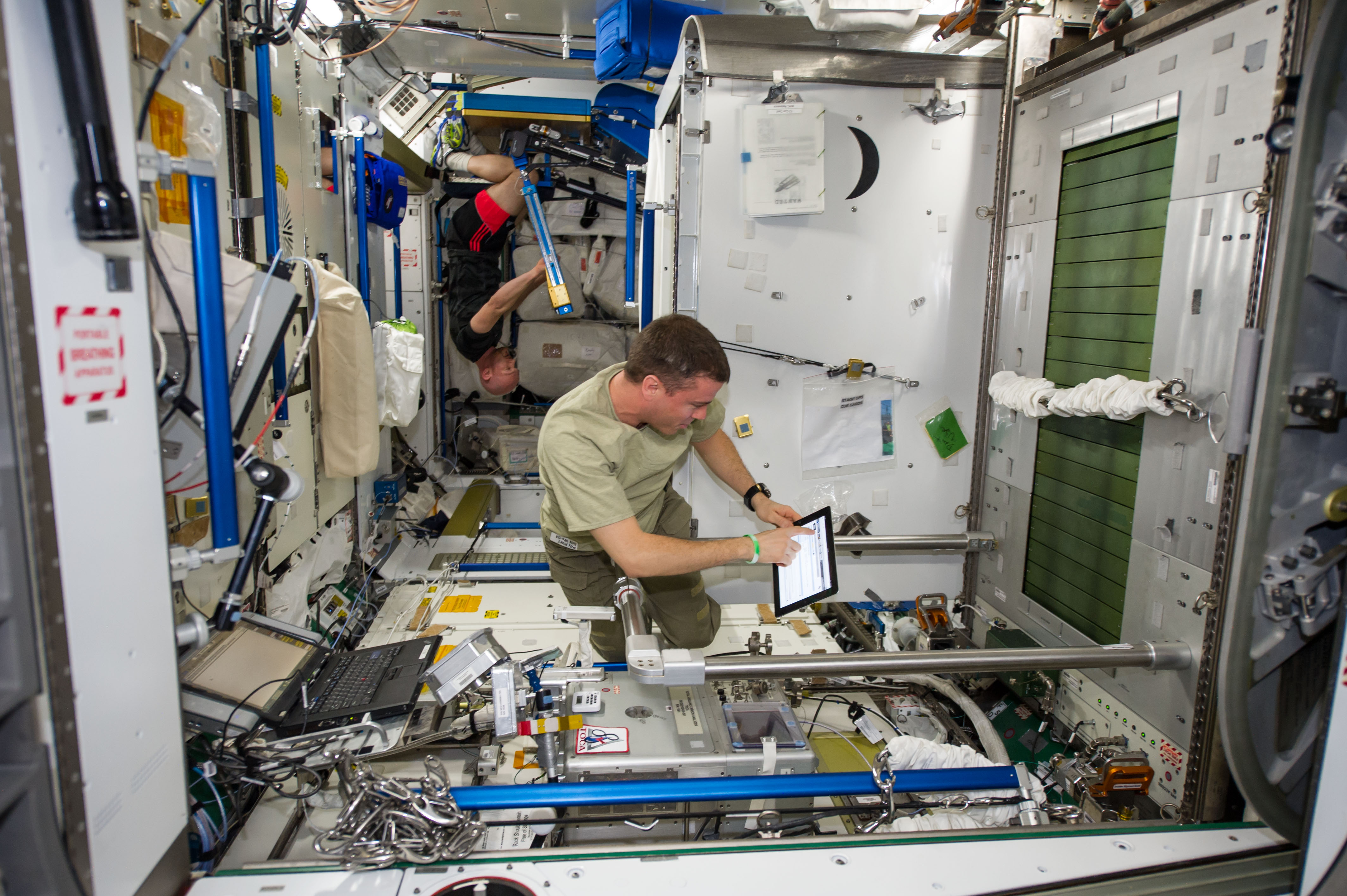}}
\caption{ \label{TOCA}  Astronauts using AMO TOCA SSC application on iPad, September 2014.}
\end{figure}

ASO's Crew Autonomy demonstrations onboard ISS required considerable systems integration.  Fortunately, most of the applications were run on laptop computers and mobile devices (see Figure \ref{TOCA}) used by ISS crew, and served by Linux and Windows servers.  Containerization and performance considerations were still important, and the operating systems, packages, and supporting software were limited, complicating integration efforts.  AMO EXPRESS also required a non-trivial interface to the Timeliner software, which runs in an embedded computer onboard ISS.  These integration efforts did not specifically complicate the use of our AI applications, but did require significant engineering, testing, and safety certification effort.

Crew Autonomy demonstrations were not only designed to show that our AI algorithms can solve computationally challenging problems; they were designed to provide decision support to astronauts.  The user interfaces for each of these demonstrations were designed in conjunction with user studies conducted by human factors engineers; see \cite{kn:AMOTOCAAIMag}, \cite{kn:AaBaVaMo} for some of the results of these user interface evaluations.  The on-orbit demonstrations also provided data evaluated by human factors engineers.  Each lesson learned provides guidance and feedback for the design of these decision support tools.   Sometimes, the lessons learned were surprising.  For instance,
crews put a lot of trust into the software recommendations, and were likely to follow the software's recommendations.  A limitation with automation software is that it is only as good as the knowledge coded within it, and this knowledge is initially based upon pre-flight characterization of hardware performance; mistakes in the software will lead the user astray.
Each demonstration provides insight into which features are used, and which are not.  Simplifying an application may imply exposing information on demand only, a lesson learned from repeated ISS rack powerup and configuration demonstrations.  In other cases, an AI technology was not deemed useful; anomaly detection performed for TOCA did not prompt direct action for the crew, and was generally disregarded, so it has not been used in further demonstrations.

Vehicle Systems Management autonomy technology is integrated with flight software, which in turn interfaces directly with vehicle hardware.
The core Flight System (cFS) is a platform and project independent reusable software framework and set of reusable software applications \cite{kn:CFS}. 
There are three key aspects to the cFS architecture: a dynamic run-time environment, layered software, and a component based design. 
Autonomy applications are treated as new components and integrated with cFS via existing messages; component interaction is handled via defining new messages between components.
CFS components are integrated via a message bus; components can publish or subscribe to the message bus.  Hardware systems data (e.g. currents, pressure, temperature, etc.) is published to the bus, and received by Fault Management (ACAWS, HyDE, IMS/MMS).  Fault management, in turn, generates messages that are consumed by the executive (PLEXIL) and planner (SCIP).  PLEXIL issues commands by publishing messages that are subscribed to by the hardware.  Integration challenges include orchestration of autonomy components via messages; extraction of the right information from hardware system messages; and, last but not least, ensuring complex algorithms do not use too many resources (CPU or memory) on embedded processors. 

\section{Related Work}
NASA has previously invested in AI technologies for use in Mission Control \cite{kn:OpsTech}.
Many of the technologies developed for use in Mission Control laid the technological foundation for the autonomy work described in this paper.
Methods to mitigate the impacts of time delay on mission operations are evaluated in \cite{kn:raices}.
An investigation into crew autonomy focused on how to write procedures for astronauts to follow that reduce the need for assistance from Mission Control
\cite{kn:istar}.  This investigation did not use special technology to assist astronauts in following procedures, but focused instead on writing guidelines, formatting, and information content.  
NASA conducted a series of crew self-scheduling demonstrations
\cite{kn:CAST} using Web-based technology to allow astronauts to build a daily activity plan and then perform the activities in this plan.
Timeliner has been extensively used to automate payload operations, and some additional activities, onboard ISS
\cite{kn:HAL}.
Science-driven extra-vehicular activity (EVA) was extensively studied during the recently completed BASALT project; a discussion of EVA operations planning and execution tools
is provided in \cite{kn:BASALTEVA}.
Finally, the Astrobee free-flying robot onboard ISS is capable of path and motion planning, image recognition, and hazard avoidance, all of which are well-studied problems in AI
\cite{kn:Astrobee}.




\section{Summary}

\begin{figure*}
\centering
{\small
\begin{tabular}{|c|c|c|c|c||c|c|c|} \hline
				&SSC	 		& TOCA 		& Orion ACAWS 	& Robotics		& AMO EXPRESS 	& Habitat VSM 		& Orion VSM 		\\ \hline \hline
Monitoring 		& 161			& 22			& 2500			& 54				&236				& 208			& 5000			\\ \hline
Planning			& -				& 6/10		& -				& 70/28			& -				& 312/25			& -  				\\ \hline
Execution			& -				& -			& -				& 70				&59		 		& 312			&  -				\\ \hline
Fault Mgt			&161				& 70			& 3500			&-				& -				& 159			& 5000			\\ \hline
Anomaly Detection	& -				&17			& -				&-				& -				& 38				& -				\\ \hline \hline
User Interfaces		&\checkmark		&\checkmark	& \checkmark		&				& \checkmark		&  				& 				\\ \hline 
Flight Software		&				&			&				&				& \checkmark		& \checkmark		& \checkmark		\\ \hline 
Machine Learning	&\checkmark		&			&				&				& 				& \checkmark		& 				\\ \hline 
\end{tabular}}
\caption{ \label{UsesWithMetrics} AI and associated technology used for ASO projects.  
For Monitoring, entry indicates the number of distinct monitored parameters; for Robotics, monitored parameters equals number of states in the planning domain.
For Planning, entry x/y indicates plan size/constraints; for Robotics, constraints equals number of distinct action types.
For Fault Mgt,  entry indicates number of detectable faults. 
For Anomaly Detection, entry indicates how many parameters are used to detect anomalies.}
\end{figure*}

NASA's future exploration missions will require smart vehicles, smart robots, and decision support to aid astronauts.  NASA is investing in AI technology, both internally developed and commercial, to make this vision a reality.  Automated planning, plan execution, and fault management technology using both automated reasoning and machine learning, have all been developed and demonstrated in a variety of operational demonstrations, both onboard the ISS and in analog environments.  
Figure \ref{UsesWithMetrics} shows the range of AI and associated technologies used in each of ASO's projects (except for \cite{kn:AMO}), as well as quantitative information indicating the rough complexity
of the problem to be solved.  
These early investments will ensure the eventual AI solutions are robust and resilient to changes in environment, faults and failures, and also the somewhat unpredictable nature of human-machine collaboration.
Lessons learned on systems engineering and integration, both 'down' to the flight software, and 'up' to the crew, will help design the proper interactions both between AI components, and between the AI components and the vehicle and its crew.
With further maturation of these AI technologies, NASA will be ready to conduct its planned future human exploration missions.

\bibliography{master}
\bibliographystyle{aaai}

\end{document}